\documentclass{article}

\usepackage[preprint]{neurips_2024}

\usepackage[utf8]{inputenc} % allow utf-8 input
\usepackage[T1]{fontenc}    % use 8-bit T1 fonts
\usepackage{hyperref}       % hyperlinks
\usepackage{url}            % simple URL typesetting
\usepackage{booktabs}       % professional-quality tables
\usepackage{amsfonts}       % blackboard math symbols
\usepackage{nicefrac}       % compact symbols for 1/2, etc.
\usepackage{microtype}      % microtypography
\usepackage{xcolor}         % colors
\usepackage{graphicx}

\title{The Role of Model Architecture and Scale in Predicting Molecular Properties: Insights from Fine-Tuning RoBERTa, BART, and LLaMA}

\author{%
    Youngmin Lee\thanks{Equal Contribution}\\
    Department of Computing \& Mathematics\\
    Oral Roberts University\\
    \texttt{ttmn\_ym@oru.edu} \\
  \And
  Andrew S.I.D. Lang\footnotemark[1] \\
  Department of Computing \& Mathematics\\
  Oral Roberts University\\
  \texttt{alang@oru.edu} \\
  \AND
  Duoduo Cai \\
  Department of Computing \& Mathematics\\
  Oral Roberts University\\
  \texttt{duoduo0321@oru.edu} \\
  \And
  Stephen R. Wheat \\
  Department of Computing \& Mathematics\\
  Oral Roberts University\\
  \texttt{swheat@oru.edu} \\
}

\begin{document}

\maketitle

\begin{abstract}
    The application of Artificial Intelligence (AI) in cheminformatics, specifically through the use of fine-tuned Large Language Models (LLMs), has shown significant potential in predicting molecular properties. This study introduces a systematic framework to compare the efficacy of LLMs for fine-tuning across various cheminformatics tasks. Employing a uniform training methodology, we assessed three well-known models—RoBERTa, BART, and LLaMA—on their ability to predict molecular properties using the Simplified Molecular Input Line Entry System (SMILES) as a universal molecular representation format. Our comparative analysis involved pre-training 18 configurations of these models, with varying parameter sizes and dataset scales, followed by fine-tuning them on six benchmarking tasks from DeepChem. We maintained consistent training environments across models to ensure reliable comparisons. This approach allowed us to assess the influence of model type, size, and training dataset size on model performance. Specifically, we found that LLaMA-based models generally offered the lowest validation loss, suggesting their superior adaptability across tasks and scales. However, we observed that absolute validation loss is not a definitive indicator of model performance - contradicts previous research - at least for fine-tuning tasks; instead, model size plays a crucial role. Through rigorous replication and validation, involving multiple training and fine-tuning cycles, our study not only delineates the strengths and limitations of each model type but also provides a robust methodology for selecting the most suitable LLM for specific cheminformatics applications. This research underscores the importance of considering model architecture and dataset characteristics in deploying AI for molecular property prediction, paving the way for more informed and effective utilization of AI in drug discovery and related fields.
\end{abstract}

\section{Introduction}
\label{intro}
The use of Artificial Intelligence (AI), including fine-tuned Large Language Models (LLMs), for predicting molecular properties has become increasingly prevalent [1]. However, the development of these models often focuses on achieving high benchmark scores and sometimes neglects how variations in training datasets, molecular representations, and training parameters influence overall performance [2]. This complicates the task of selecting the best LLM to fine-tune for a specific task. Our research presents a straightforward framework designed to compare LLMs, assessing their suitability for fine-tuning across a range of tasks. This approach is demonstrated through the evaluation of three popular open-source models for fine-tuning for molecular property prediction: RoBERTa, a BERT (Bidirectional Encoder Representations from Transformers)-based model [3], BART (Bidirectional and Auto-Regressive Transformers) [4], and LLaMA (Large Language model Meta AI) [5], chosen for their proven utility in chemistry-specific Natural Language Processing (NLP) applications. The intent of this study is not to identify the universally best LLM across all models and tasks but to demonstrate an effective method for comparing various LLMs. In conducting our comparative analysis, we adopted a consistent training methodology and used uniform training data across all models to fine-tune RoBERTa, BART, and LLaMA for molecular property prediction tasks. Additionally, given the extensive use of the Simplified Molecular Input Line Entry System (SMILES) for representing chemical structures in the fine-tuning process of LLMs [6-11], our methodology employs SMILES encoding as the standard form of molecular representation. While acknowledging the existence of alternative molecular encoding techniques [12], our study primarily aims to showcase a comparative methodology for determining the most effective LLM for a specific task under uniform training conditions.

\section{Related Works}
\label{related-works}
Recent studies have highlighted the utility of fine-tuning BERT and its variants to create specialized models such as CHEM-BERT [6] and ChemBERTa-2 [7]. These models can undergo further fine-tuning for molecular property prediction [8], indicating the versatility of BERT-based architectures. Similarly, research has demonstrated the usefulness of BART-based architectures for small-molecule drug discovery tasks, exemplified by models such as Chemformer [8] and MegaMolBART [9], with the latter also being applied to molecular property prediction tasks [10]. Furthermore, LLaMA (and other recently released open models) have also been fine-tuned for chemistry-related tasks through instruction tuning [13], suggesting their potential use in various chemistry applications. Finally, Molformer, leveraging techniques like encoder-based architecture, Masked Language Modeling (MLM), linear attention, and bucketing, is prominently used in cheminformatics for predictive tasks, including molecular property prediction and chemical reaction outcomes. Its capability for de novo drug design allows for the generation of novel molecular structures with desired properties, aiding in efficient drug discovery processes. Additionally, Molformer excels in tasks such as molecular similarity assessment and clustering, enhancing the screening and optimization of compounds in large datasets [14].

These recent advancements in fine-tuning BART, RoBERTa, and LLaMA (amongst others) underscore the importance of foundational LLMs in addressing domain-specific challenges in chemistry and illustrate the need to have the ability to directly compare the performance of such models on the same task.

\section{Methodology}
\label{methodology}

\subsection{Pre-training Models for Multi-Task Regression (MTR)}
\label{mtr}
To assess the effects of model size and dataset size on the performance of foundational transformer models (RoBERTa, BART, and LLaMA), the authors established 18 configurations by training two versions of each model—one with 13 million parameters and another with 30 million parameters—across three distinct training dataset sizes: 10 million, 20 million, and 30 million instances. This setup was designed to explore scalability and performance impacts. To ensure experimental consistency, the same data sets were used for each dataset size, with fixed random seeds to guarantee reproducibility (see Table~\ref{table-comb}).

\begin{table}
  \caption{Model combination chart: The 18 experimental configurations result from the product of three variables: three foundational model types, two parameter sizes, and three training dataset sizes.}
  \label{table-comb}
  \centering
  % \small % Reduce font size
  \footnotesize
  \setlength{\tabcolsep}{4pt} % Reduce column separation
  \renewcommand{\arraystretch}{0.9} % Reduce vertical spacing
  \begin{tabular}{ll}
    \toprule
    % \multicolumn{2}{c}{Part}                   \\
    % \cmidrule(r){1-2}
    \multicolumn{1}{c}{Factor} & \multicolumn{1}{c}{Values} \\
    \midrule
    Model Type (MT) & \multicolumn{1}{c}{ChemBART | ChemBERTa | ChemLLaMA} \\
    Model Size (MS) & \multicolumn{1}{c}{Small (13M) \ \ \ \  | \  \ \ \  Medium (30M)} \\
    Data Size  (DS) & \multicolumn{1}{c}{10M \ \ \ \  | \ \ \  \ 20M \ \ \ \   | \ \ \ \ 30M \ \ } \\
    \bottomrule
  \end{tabular}
\end{table}

\subsubsection{Dataset}
\label{mtr-dataset}
We downloaded the canonicalized SMILES dataset utilized for Molformer [14], initially processing 33 million SMILES strings. After removing duplicates, we randomly selected 30 million entries for further analysis. Using DeepChem’s RDKitDescriptors Featurizer, we calculated descriptors for each SMILES string [15, 16]. This process revealed significant skewing in some property columns due to a few extreme values (e.g., ‘Kappa3’ values exceeding 1e+36, while most are below 1e+3). We addressed this by removing outliers and normalizing all properties using z-scores. Finally, we excluded any columns where the minimum and median values were identical, resulting in a curated initial training dataset of 30 million distinct compounds (30M rows) with 105 properties (106 columns: 1 identifier + 105 properties). To set the uniform training environment, we used the exact same training and validation data for all transformer models for each data size. Note that we shuffle the order of training data for each epoch.

\subsubsection{Tokenizer}
\label{mtr-tokenizer}
All tokenizers for each transformer model are based on the tokenizer from ChemBERTa [7]. The authors have replaced the special tokens with the default settings from Hugging Face. The maximum sequence length for the tokenizers is 512. Note that the ChemLLaMA tokenizer adds the `eos' tokens at the end of the sequence before padding, while the default LLaMA tokenizer does not.

\subsubsection{Model for MTR (pre-trained)}
\label{mtr-model}
\begin{figure}
  \centering
  % \fbox{\rule[-.5cm]{0cm}{4cm} \rule[-.5cm]{4cm}{0cm}}
  \includegraphics[width=10cm]{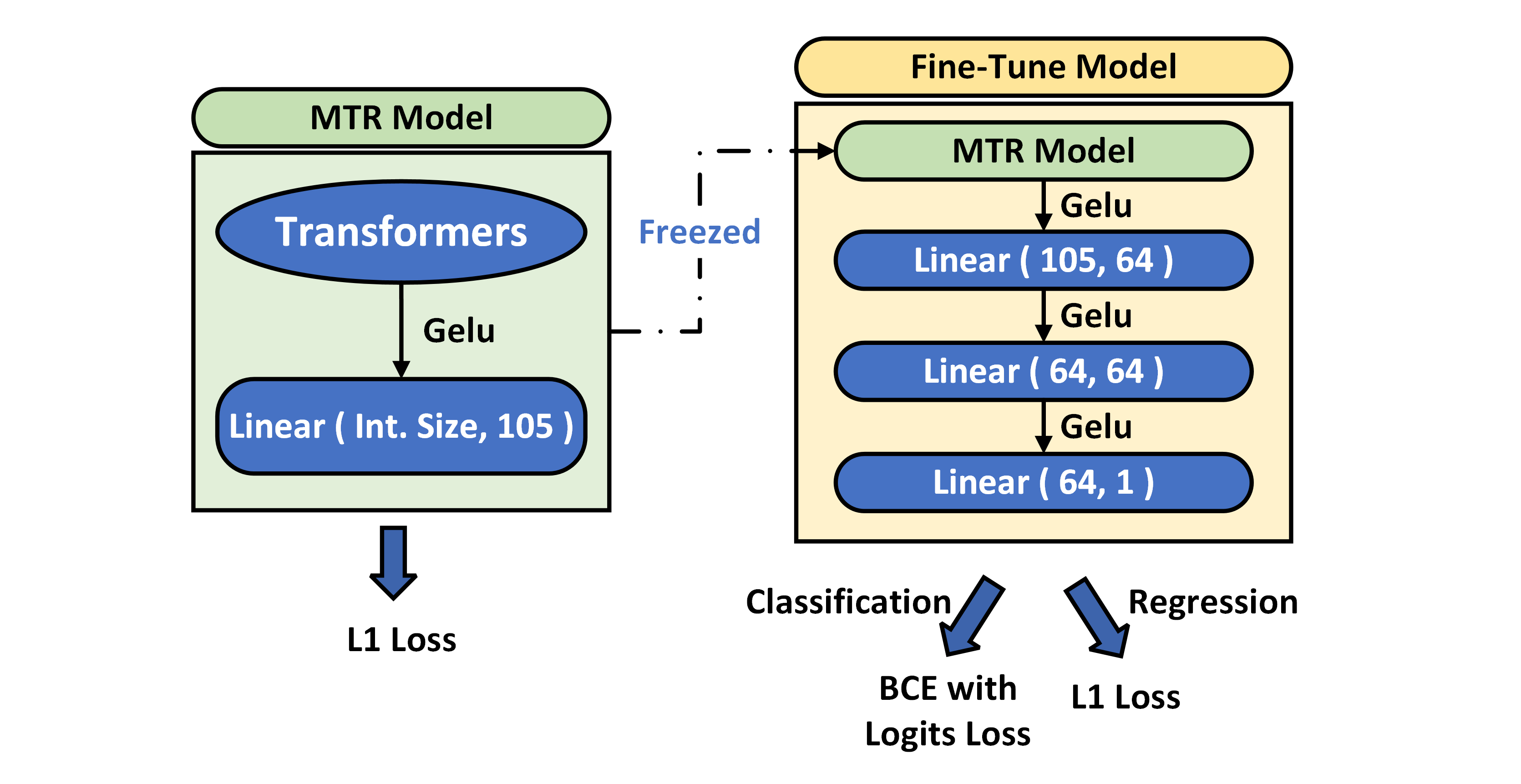}
  \caption{The Architecture of Multi-Task Regression model and Fine-Tuned model.}
  \label{fig-architecture-mtr}
\end{figure}

The authors utilized three types of transformer-based models, representing encoder-based, decoder-based, and encoder-decoder-based architectures: RoBERTa, LLaMA, and BART, respectively. Default settings were used for each model, with adjustments made only to the special token IDs and vocabulary size, as illustrated in Figure~\ref{fig-architecture-mtr}. The models were trained with two parameter sizes, featuring 30 million (Medium) and 13 million (Small) parameters as calculated using the .num\_param() method from the Hugging Face `transformers' library. During the training, BART and LLaMA models calculated loss using the `eos' token, whereas RoBERTa utilized the `bos' token. We employed L1Loss, corresponding to Mean Absolute Error (MAE). The MTR model families discussed in this paper—ChemBERTa, ChemLLaMA, and ChemBART, see Table~\ref{table-model-size}—correlate with the architectures of RoBERTa, LLaMA, and BART, respectively.

\begin{table}
  \caption{Model Size Chart: The configuration of each transformer model (Model Type) by Model Size. All other parameters are the default values provided for each model on Hugging Face.}
  \label{table-model-size}
  \centering
  % \small % Reduce font size
  \footnotesize
  \setlength{\tabcolsep}{4pt} % Reduce column separation
  \renewcommand{\arraystretch}{0.9} % Reduce vertical spacing
  \begin{tabular}{clcccc}
    \toprule
    \multicolumn{2}{c}{MTR Model} \\
    \cmidrule(lr){1-2}
    \multicolumn{1}{c}{Size} & \multicolumn{1}{c}{Type} & Hid. Size & Int. Size & Hid. Layers & Att. Heads \\
    \midrule
           & ChemBART  & 624 | 624 & 624 | 624 & 2 | 2 & 2 | 2 \\
    Small  & ChemBERTa & 620 & 710 & 5 & 5 \\
    (13M)  & ChemLLaMA & 600 & 620 & 5 & 5 \\
    % \cmidrule{2-6}
    \midrule
           & ChemBART  & 768 | 768 & 768 | 768 & 3 | 3 & 3 |3 \\
    Medium & ChemBERTa & 768 & 768 & 8 & 8 \\
    (30M)  & ChemLLaMA & 768 & 768 & 7 & 8 \\
    \bottomrule
  \end{tabular}
 
\end{table}

\subsubsection{Training}
\label{mtr-training}
We configured both the training and validation phases with a batch size of 64. For optimization, we employed the AdamW optimizer, using its default settings. We also used the `Linear Warmup Cosine Annealing Learning Rate' scheduler from Pytorch Lightning Bolts, which gradually ramped up the learning rate to a peak of 0.0001 by the end of the first epoch (starting from epoch 0) and sustained this rate across a total of 7 epochs. Our MTR models were trained on four NVIDIA A30 GPUs using a Distributed Data Parallelism (DDP) strategy, facilitated by the Pytorch Lightning framework.

\subsection{Models for Fine-tuning (FT)}
\label{ft}

\subsubsection{Dataset \& Tokenizer}
\label{ft-dataset-tokenizer}
We selected six benchmarking tasks from DeepChem to evaluate the performance of our transformer models after fine-tuning. These tasks, detailed in Table~\ref{table-ft-dataset}, are representative of common benchmark challenges in cheminformatics. For regression analyses, we utilized the `bace\_regression,' `delaney,' and `lipo' datasets. For classification tasks, we selected `bace\_classification,' `hiv,' and `tox21\_sr\_p53.' Consistency in preprocessing was maintained by using the same tokenizers as those applied in the pre-training phase of our models.

\begin{table}
  \caption{Fine-tune dataset: Datasets employed for model Fine-tuning - task classification and dataset sizes.}
  \label{table-ft-dataset}
  \centering
  % \small % Reduce font size
  \footnotesize
  \setlength{\tabcolsep}{4pt} % Reduce column separation
  \renewcommand{\arraystretch}{0.9} % Reduce vertical spacing
  \begin{tabular}{ccccccc}
    \toprule
    % \multicolumn{2}{c}{Part}                   \\
    % \cmidrule(r){1-2}
    \multicolumn{1}{c}{Fine-tune Dataset} & \multicolumn{3}{c}{Regression} & \multicolumn{3}{c}{Classification} \\
    % \cmidrule{1}
    \cmidrule(l){1-1} \cmidrule(lr){2-4} \cmidrule(r){5-7}
    Name & Bace & Delaney & Lipo & Bace & Hiv   & Tox21 \\
    \midrule
    Size & 1513 & 1127    & 4200 & 1513 & 40000 & 8000 \\
    \bottomrule
  \end{tabular}
\end{table}

\subsubsection{Models Used for Fine-Tuning}
\label{ft-model}
In the fine-tuning process, we utilized the 18 pre-trained models from our MTR series. These models were initially frozen to preserve the learned features and then appended with linear layers incorporating the Gelu activation function, as depicted in Figure~\ref{fig-architecture-mtr}. For the loss functions, the fine-tuned models employed Mean Absolute Error (MAE, or L1 Loss) for regression tasks and Binary Cross-Entropy with Logits (BCE with Logits Loss) for classification tasks.

\subsubsection{Training}
\label{ft-training}
During the fine-tuning of our models, we allocated one GPU per model due to the relatively smaller size of the fine-tuning datasets compared to those used for pre-training. Consequently, we did not employ parallel processing techniques and instead adhered to the default training settings. We retained the same learning rate scheduler as in the pre-training phase but increased the peak learning rate from 0.0001 to 0.01 for both regression and classification tasks to better suit the scale of the datasets. All other training configurations were maintained as in the initial setup. To ensure the reliability of our results and reduce potential biases from single-model evaluations, we repeated the fine-tuning process five times. This approach enabled us to compile and analyze statistics across multiple iterations, enhancing the robustness of our findings.

    \section{Results \& Discussion}
\label{result}

\subsection{Model Multi-Task Regression}
\label{result-model-mtr}

\subsubsection{Training and Validation Loss}
\label{result-mtr-loss}
\begin{figure}
  \centering
  % \fbox{\rule[-.5cm]{0cm}{4cm} \rule[-.5cm]{4cm}{0cm}}
  \includegraphics[width=13.97cm]{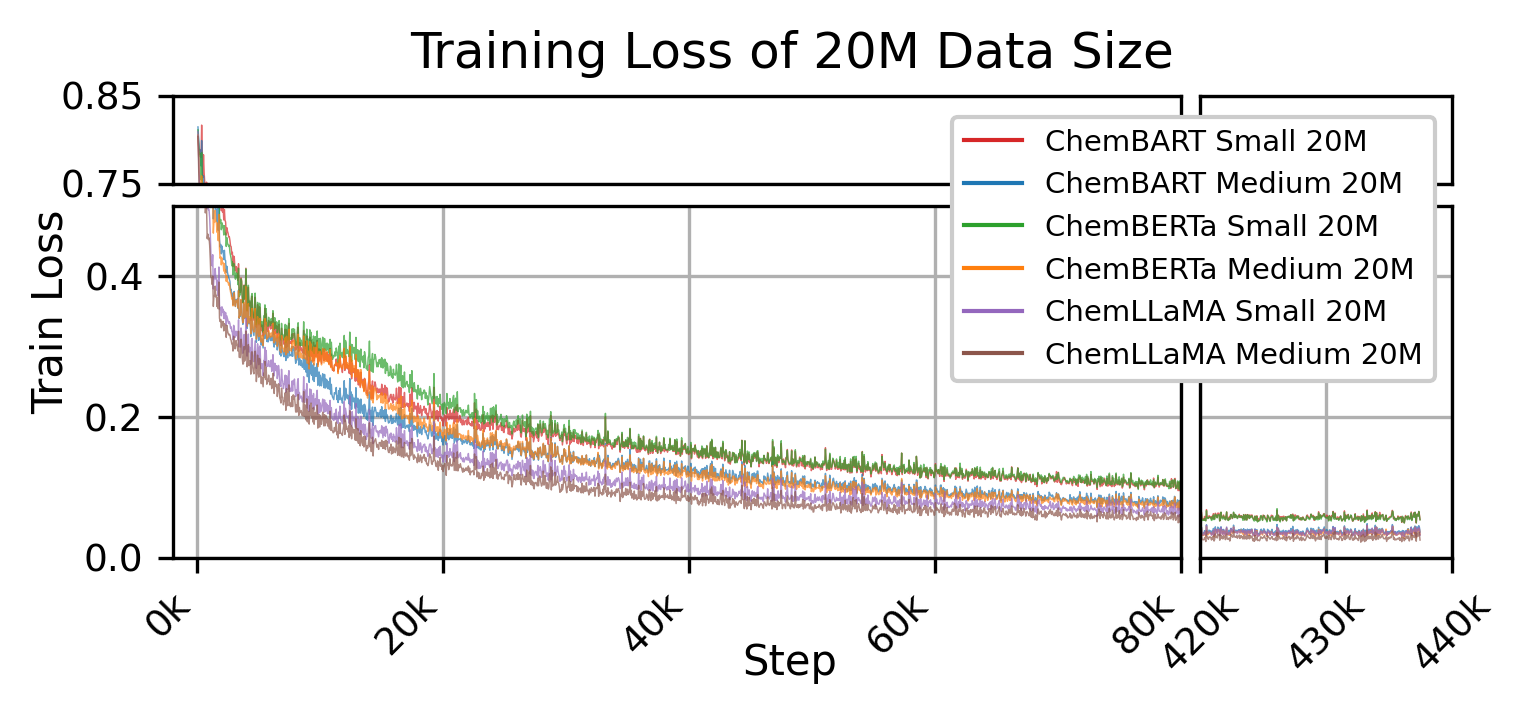}
  \caption{Training Loss vs. Steps for Each Model Type by Model Size with a Data Size of 20 millions. The loss trends for 10 million and 20 million datasets are similar.}
  \label{fig-loss-20m}
\end{figure}

\begin{figure}
  \centering
  \includegraphics[width=13.97cm]{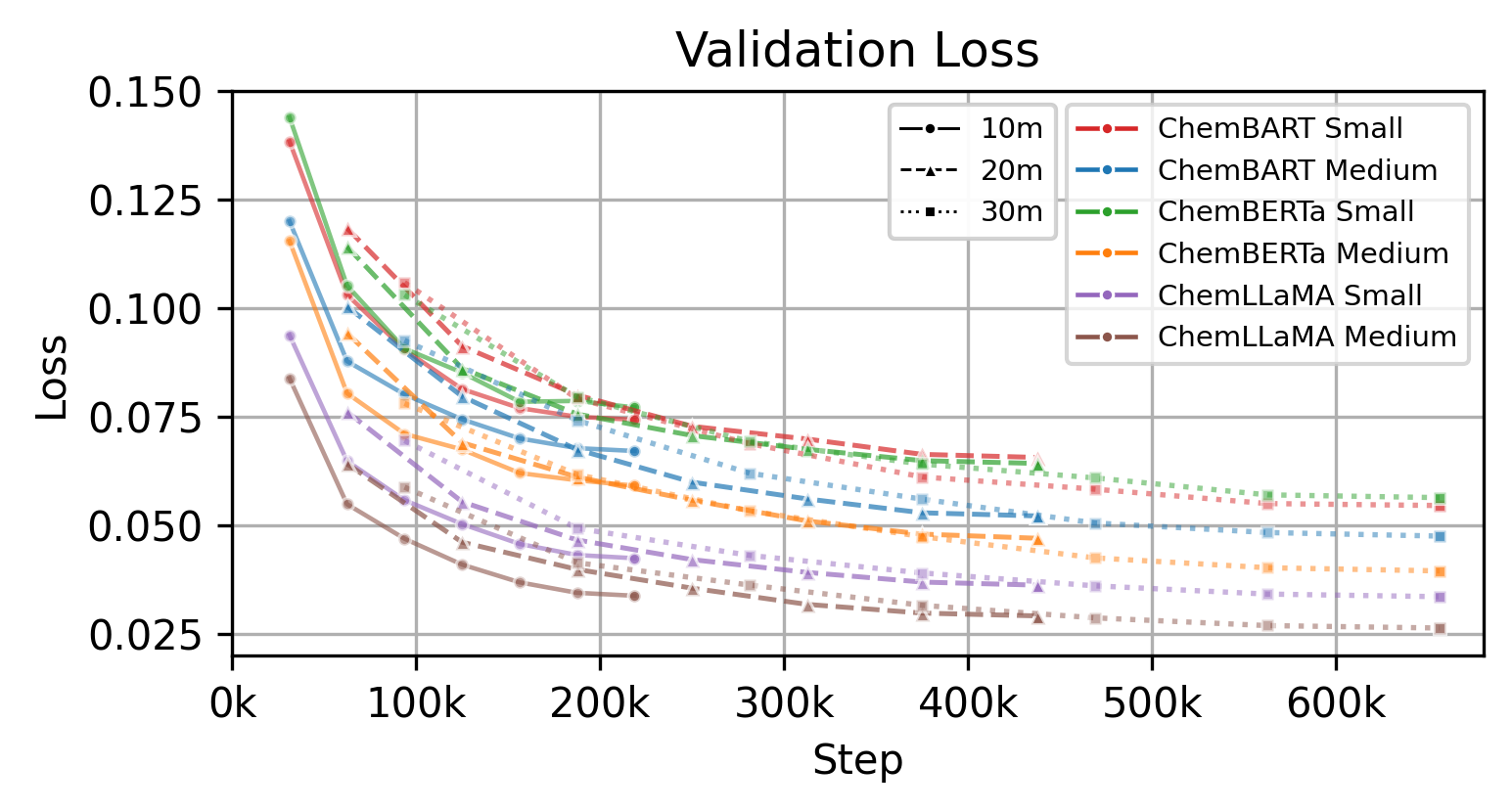}
  \caption{Validation Loss vs. Steps for Each Model Type Across All Model Sizes and Data Sizes (10, 20, and 30 million).}
  \label{fig-vloss}
\end{figure}

During the training of the MTR models, we observed that BART and RoBERTa-based models exhibited similar learning rate behaviors across all model sizes. However, ChemBART's learning rate decreased more rapidly than ChemBERTa's at the outset, across all sizes, as shown in Figure~\ref{fig-loss-20m} (data for 20 million parameters displayed, with other sizes presenting similar trends). Regarding validation loss, the smaller-sized models of ChemBART and ChemBERTa displayed comparable loss values across various data sizes and epochs. However, for the medium-sized models, ChemBART consistently showed lower validation losses in all cases, as illustrated in Figure~\ref{fig-vloss}. ChemLLaMA achieved the lowest validation loss values, irrespective of epoch or model size.

Additionally, we confirmed that larger MTR models tend to reach saturation more quickly than their smaller counterparts, aligning with previous findings [17]. Interestingly, as Figure~\ref{fig-vloss} illustrates, ChemLLaMA models of the same size exhibited lower validation losses with smaller datasets compared to larger ones throughout the gradient steps. Nonetheless, models trained with larger datasets ultimately achieved lower validation losses as the number of gradient steps increased.

\subsection{Model Fine-Tuning}
\label{result-ft}
The authors trained each fine-tuned model for seven epochs and replicated this process five times to mitigate potential statistical anomalies from any single run. The total count of metrics (MAE/BCE) recorded for each finely-tuned (FT) dataset, encompassing all model types (MT), model sizes (MS), and data sizes (DS), amounted to 3 $\times$ 2 $\times$ 3 $\times$ 7 (epochs for MTR models) $\times$ 7 (epochs for FT models) $\times$ 5 (iterations) = 4,410 performance values. The `best' metric over all epochs was determined by selecting the model with the lowest validation loss and recorded in a `best metrics set' for 3 (MT) $\times$ 2 (MS) $\times$ 3 (DS) $\times$ 5 (iterations) = 90 models. From these 90 models, we calculate the performance metric (RMSE/AUC) using the test dataset.

For the training, we employed Mean Absolute Error (MAE) for regression and Binary Cross Entropy (BCE) for classification. However, for benchmarking each fine-tuning task with the test datasets, we used Root Mean Squared Error (RMSE) and the Area Under the Curve (AUC) of the Receiver Operating Characteristics (ROC).

\subsubsection{Total Error Sum (TES) and its Standard Deviation (STD)}
\label{result-ft-es-std}
We assess the performance of various finely-tuned models across multiple tasks using the following categories: Model Type, Model Type \& Model Size, Model Type \& Data Size, and Model Size \& Data Size. Our evaluation process is as follows:

\begin{enumerate}

    \item {We compute the average RMSE (for regression tasks) or AUC (for classification tasks) for each group: Model Type, Model Type \& Model Size, Model Type \& Data Size, and Model Size \& Data Size.}
    \item {We establish the optimal performance — the lowest average RMSE in regression tasks or the highest average AUC in classification tasks — as the benchmark for each group.}
    \item {For each individual model run, we measure the deviation of the RMSE or AUC from the established benchmark. These deviations are signed to reflect that a lower RMSE or higher AUC is preferable, aligning with performance standards for both regression and classification tasks.}
    \label{step3}
    \item {We then aggregate these deviations, grouped by the aforementioned categories, into an Error Sum (ES) for each task.}
    \item {We calculate the Total Error Sum (TES) by summing all the ES values, providing a comprehensive performance metric across tasks. Additionally, we determine the Standard Deviation (STD) based on the deviations from the benchmark identified in Step~\ref{step3}.}

\end{enumerate}

This structured approach enables a detailed model performance analysis relative to established standards across different configurations.

\subsubsection{Overall Test Evaluation}
\label{result-ft-overall}
Table~\ref{table-best-mt} displays the average best metrics (RMSE/AUC) for our finely-tuned models, grouped by model type. For example, the average of the best metric for the BACE benchmark task is calculated from 30 RMSE values, derived from combinations of 2 model sizes, 3 data sizes, and 5 runs. This yields three comparative averages—one for each model type. Among them, ChemBART records the lowest average RMSE at 0.896. Average metrics for other tasks are calculated similarly. Table~\ref{table-best-mtms} splits the results by model size.

\begin{table}[!htb]
    \begin{minipage}{.6\linewidth}
      \caption{Best Avg. Metrics by MT}
      \label{table-best-mt}
      \centering
      % \small % Reduce font size
      \footnotesize
      \setlength{\tabcolsep}{3pt} % Reduce column separation
      \renewcommand{\arraystretch}{0.9} % Reduce vertical spacing
        \begin{tabular}{llcl}
            \toprule
            \multicolumn{2}{c}{FT Dataset}  \\
            \cmidrule(lr){1-2}
            \multicolumn{1}{c}{Tasks} & Name & Avg. Metrics & \multicolumn{1}{c}{MT} \\
            \midrule
                                           & Bace    & 0.895847 & ChemBart    \\
            \multicolumn{1}{c}{Regression} & Delaney & 0.478322 & ChemBart    \\
            \multicolumn{1}{c}{(RMSE)}       & Lipo    & 0.735585 & ChemBerta \\
            \midrule
                                               & Bace    & 0.802156 & ChemBerta  \\
            \multicolumn{1}{c}{Classification} & Hiv     & 0.769427 & ChemLlama  \\
            \multicolumn{1}{c}{(AUC-ROC)}        & Tox21   & 0.789914 & ChemLlama  \\
            \bottomrule
        \end{tabular}
    \end{minipage}%
    \begin{minipage}{.4\linewidth}
      \centering
        \caption{Best Avg. Metrics by MTMS}
        \label{table-best-mtms}
        % \small % Reduce font size
        \footnotesize
        \setlength{\tabcolsep}{3pt} % Reduce column separation
        \renewcommand{\arraystretch}{0.9} % Reduce vertical spacing
        \begin{tabular}{cl}
            \toprule
            \multicolumn{2}{c}{Sharing FT Dataset}  \\
            \cmidrule(lr){1-2}
            \multicolumn{1}{c}{Avg. Metrics} & \multicolumn{1}{c}{MTMS} \\
            \midrule
            0.867828 & ChemBART Medium    \\
            0.452183 & ChemLLaMA Medium    \\
            0.731018	 & ChemBERTa Medium \\
            \midrule
            0.802258 & ChemBERTa Small \\
            0.774683 & ChemLLaMA Small  \\
            0.794821 & ChemBART Small  \\
            \bottomrule
        \end{tabular}
    \end{minipage} 
\end{table}

\subsubsection{Best Metrics Grouped by Model Type and Model Size (MTMS)}
\label{result-ft-grouped-mtms}
When focusing solely on the transformer's Model Type and disregarding the size, ChemBART outperforms the other models in regression tasks but ranks lowest in classification tasks. Whereas, ChemLLaMA exhibits the worst performance (highest TES) in regression tasks across both sizes, as shown in Table~\ref{table-tes-mt}. However, ChemLLaMA medium has the best overall performance (lowest TES values) among all Model Types for medium-sized models in both regression and classification tasks, as highlighted in Table~\ref{table-tes-mtms}. These tables categorize the best metric sets from each dataset, highlighting differences across model configurations.

\begin{table}[!htb]
    \begin{minipage}[t]{.45\linewidth}
      \caption{TES and STD grouped by MT}
      \label{table-tes-mt}
      \centering
      % \small % Reduce font size
      % \footnotesize
      \scriptsize
      \setlength{\tabcolsep}{3pt} % Reduce column separation
      \renewcommand{\arraystretch}{0.9} % Reduce vertical spacing
      \begin{tabular}{lllll}
        \toprule
        \multicolumn{1}{c}{Group} & \multicolumn{2}{c}{Regression} & \multicolumn{2}{c}{Classification} \\
        \cmidrule(lr){1-1} \cmidrule(lr){2-3} \cmidrule(lr){4-5}
        \multicolumn{1}{c}{MT}  & \multicolumn{1}{c}{TES} & \multicolumn{1}{c}{STD} & \multicolumn{1}{c}{TES} & \multicolumn{1}{c}{STD} \\
        \midrule
        ChemBART  & \textbf{0.001455} & \textit{0.055811} & 0.017267 & 0.020405  \\
        ChemBERTa & 0.036971 & 0.072974 & 0.012621 & \textit{0.019049}  \\
        ChemLLaMA & 0.042069 & 0.074816 & \textbf{0.010000} & 0.021938  \\
        \bottomrule
      \end{tabular}
    \end{minipage}%
    \begin{minipage}[t]{.55\linewidth}
      \centering
        \caption{TES and STD grouped by MTMS}
        \label{table-tes-mtms}
        % \small % Reduce font size
        % \footnotesize
        \scriptsize
        \setlength{\tabcolsep}{3pt} % Reduce column separation
        \renewcommand{\arraystretch}{0.9} % Reduce vertical spacing
          \begin{tabular}{llllll}
            \toprule
            \multicolumn{2}{c}{Group} & \multicolumn{2}{c}{Regression} & \multicolumn{2}{c}{Classification}  \\
            \cmidrule(lr){1-2} \cmidrule(lr){3-4} \cmidrule(lr){5-6}
            \multicolumn{1}{c}{MT}  & \multicolumn{1}{c}{MS} & \multicolumn{1}{c}{TES} & \multicolumn{1}{c}{STD} & \multicolumn{1}{c}{TES} & \multicolumn{1}{c}{STD}\\
            \midrule
            ChemBART & Small & 0.085963 & 0.062398 & 0.028660 & 0.024561 \\
                     & Medium & 0.034399 & 0.048846 & 0.026405 & \textit{0.015154} \\
            \cmidrule(r){1-4} \cmidrule(l){5-6}
            ChemBERTa & Small & 0.086802 & 0.065119 & 0.023367 & 0.019636 \\
                      & Medium & 0.104591 & 0.083724 & 0.022404 & 0.019529 \\
            \cmidrule(r){1-4} \cmidrule(l){5-6}
            ChemLLaMA & Small & 0.171555 & 0.094720 & 0.022002 & 0.023530 \\
                      & Medium & \textbf{0.030035} & \textit{0.039855} & \textbf{0.018527} & 0.019365 \\
            \bottomrule
          \end{tabular}
    \end{minipage} 
\end{table}

The baseline values for each dataset are taken from Table~\ref{table-best-mtms}. Additionally, the best metrics are organized by MTMS from the original best metric set for each dataset.

As previously mentioned, ChemLLaMA with a 'Medium' model size demonstrates the best performance across all regression and classification tasks, even though ChemLLaMA ‘Small’ has the worst performance in regression tasks among all MTMS combinations. The 'Medium' size ChemBART also performs impressively in regression tasks.

Furthermore, our analysis reveals that the performance of each Model Type generally scales with Model Size across all task types, with the notable exception of ChemBERTa in regression tasks.

\subsubsection{Best Metrics Grouped by Model Type/Data Size (MTDS) and Model Size/Data Size (MSDS)}
\label{result-ft-grouped-mtds-msds}
When the average metrics and TES are calculated from the best model and grouped by Model Type and Data Size (MTDS), as presented in Tables~\ref{table-best-mtds} and~\ref{table-tes-mtds}, ChemBART, utilizing 10M of MTR data, excels in regression tasks compared to other groups. For classification tasks, ChemLLaMA with 30M of data achieves the lowest TES value. ChemBERTa with 30M of data also displays strong performance in classification, alongside other groups with similarly low TES values. Additionally, performance for ChemBERTa in regression tasks and for ChemLLaMA in classification tasks shows a clear proportional relationship to data size.

Considering Model Size and Model Type, as detailed in Tables~\ref{table-best-msds} and~\ref{table-tes-msds}, a Medium Model Size with 10M Data Size yields the best (lowest) TES value in regression tasks. Generally, larger Model Sizes outperform smaller ones across all Data Sizes. In classification tasks, Small Model Sizes with 30M Data Size achieve the lowest TES. However, with the exception of the 30M data scenarios, Medium Model Sizes tend to exhibit lower errors than smaller sizes. Notably, in regression tasks, Small Model Sizes demonstrate a non-proportional relationship between TES and Data Size. Conversely, in classification tasks, there is a consistent proportional relationship between performance and Data Size, regardless of Model Size.

\begin{table}[!htb]
    \begin{minipage}{.6\linewidth}
      \caption{Best Avg. Metrics by MTDS}
      \label{table-best-mtds}
      \centering
      % \small % Reduce font size
      \footnotesize
      \setlength{\tabcolsep}{3pt} % Reduce column separation
      \renewcommand{\arraystretch}{0.9} % Reduce vertical spacing
      \begin{tabular}{llcl}
        \toprule
        \multicolumn{2}{c}{FT Dataset}  \\
        \cmidrule(lr){1-2}
        \multicolumn{1}{c}{Tasks}    & Name   & \multicolumn{1}{c}{Avg. Metrics} & \multicolumn{1}{c}{MTDS} \\
        \midrule
                                       & Bace    & 0.877418 & ChemBart 10m    \\
        \multicolumn{1}{c}{Regression} & Delaney & 0.459106 & ChemBart 10m    \\
        \multicolumn{1}{c}{(RMSE)}     & Lipo    & 0.733709 & ChemBerta 20m \\
        \midrule
                                           & Bace    & 0.806304 & ChemBerta 30m \\
        \multicolumn{1}{c}{Classification} & Hiv     & 0.778740 & ChemLlama 30m  \\
        \multicolumn{1}{c}{(AUC-ROC)}      & Tox21   & 0.798990 & ChemLlama 30m  \\
        \bottomrule
      \end{tabular}
    \end{minipage}%
    \begin{minipage}{.4\linewidth}
      \centering
        \caption{Best Avg. Metrics by MSDS}
        \label{table-best-msds}
        % \small % Reduce font size
        \footnotesize
        \setlength{\tabcolsep}{3pt} % Reduce column separation
        \renewcommand{\arraystretch}{0.9} % Reduce vertical spacing
        \begin{tabular}{cl}
            \toprule
            \multicolumn{2}{c}{Sharing FT Dataset}  \\
            \cmidrule(lr){1-2}
            \multicolumn{1}{c}{Avg. Metrics} & \multicolumn{1}{c}{MSDS} \\
            \midrule
            0.879725 & Medium 10m    \\
            0.470251 & Medium 30m    \\
            0.730185 & Medium 10m \\
            \midrule
            0.801208 & Medium 20m \\
            0.773886 & Small 30m  \\
            0.796570 & Small 30m  \\
            \bottomrule
      \end{tabular}
    \end{minipage} 
\end{table}

\begin{table}[!htb]
    \begin{minipage}[t]{.5\linewidth}
      \caption{TES and STD grouped by MTDS}
      \label{table-tes-mtds}
      \centering
      % \small % Reduce font size
      % \footnotesize
      \scriptsize
      \setlength{\tabcolsep}{3pt} % Reduce column separation
      \renewcommand{\arraystretch}{0.9} % Reduce vertical spacing
      \begin{tabular}{llllll}
        \toprule
        \multicolumn{2}{c}{Group} & \multicolumn{2}{c}{Regression} & \multicolumn{2}{c}{Classification} \\
        \cmidrule(lr){1-2} \cmidrule(lr){3-4} \cmidrule(lr){5-6}
        \multicolumn{1}{c}{MT}  & \multicolumn{1}{c}{DS} & \multicolumn{1}{c}{TES} & \multicolumn{1}{c}{STD} & \multicolumn{1}{c}{TES} & \multicolumn{1}{c}{STD} \\
        \midrule
                  & \multicolumn{1}{c}{10 M} & \textbf{0.002017} & 0.048120 & 0.040099 & 0.018766 \\
        ChemBART  & \multicolumn{1}{c}{20 M} & 0.090277 & 0.067623 & 0.037477 & 0.021300 \\
                  & \multicolumn{1}{c}{30 M} & 0.030635 & \textit{0.046525} & 0.041837 & 0.021029 \\
        \cmidrule(r){1-4} \cmidrule(l){5-6}
                  & \multicolumn{1}{c}{10 M} & 0.059857 & 0.061768 & 0.046010 & 0.021386 \\
        ChemBERTa & \multicolumn{1}{c}{20 M} & 0.074375 & 0.098413 & 0.044872 & 0.020028 \\
                  & \multicolumn{1}{c}{30 M} & 0.095244 & 0.054786 & 0.014592 & \textit{0.014878} \\
        \cmidrule(r){1-4} \cmidrule(l){5-6}
                  & \multicolumn{1}{c}{10 M} & 0.081123 & 0.073842 & 0.048363 & 0.022265 \\
        ChemLLaMA & \multicolumn{1}{c}{20 M} & 0.079537 & 0.087679 & 0.036204 & 0.024237 \\
                  & \multicolumn{1}{c}{30 M} & 0.084112 & 0.064089 & \textbf{0.013043} & 0.015452 \\
        \bottomrule
      \end{tabular}
    \end{minipage}%
    \begin{minipage}[t]{.5\linewidth}
      \centering
        \caption{TES and STD grouped by MSDS}
        \label{table-tes-msds}
        % \small % Reduce font size
        % \footnotesize
        \scriptsize
        \setlength{\tabcolsep}{3pt} % Reduce column separation
        \renewcommand{\arraystretch}{0.9} % Reduce vertical spacing
        \begin{tabular}{lcllll}
            \toprule
            \multicolumn{2}{c}{Group} & \multicolumn{2}{c}{Regression} & \multicolumn{2}{c}{Classification}  \\
            \cmidrule(lr){1-2} \cmidrule(lr){3-4} \cmidrule(lr){5-6}
            \multicolumn{1}{c}{MS}  & \multicolumn{1}{c}{DS} & \multicolumn{1}{c}{TES} & \multicolumn{1}{c}{STD} & \multicolumn{1}{c}{TES} & \multicolumn{1}{c}{STD} \\
            \midrule
                       & 10 M & 0.072341 & 0.076212 & 0.033915 & 0.022837 \\
            Small      & 20 M & 0.089094 & 0.086085 & 0.029663 & 0.024573 \\
                       & 30 M & 0.095489 & 0.062148 & \textbf{0.010157} & 0.019240 \\
            \cmidrule(r){1-4} \cmidrule(l){5-6}
                       & 10 M & \textbf{0.003135} & \textit{0.042831} & 0.030993 & 0.018664 \\
             Medium    & 20 M & 0.053843 & 0.083541 & 0.024632 & 0.019120 \\
                       & 30 M & 0.024651 & 0.045687 & 0.011417 & 0.016060 \\
            \bottomrule
          \end{tabular}
    \end{minipage}% 
\end{table}

\section{Conclusion}
\label{conclusion}
In our MTR training sessions, ChemLLaMA consistently demonstrated the lowest validation loss across all model sizes and epochs. However, we observed that absolute validation loss is not a definitive indicator of model performance - at least for fine-tuning tasks; instead, model size plays a crucial role. Overall, ChemBART excels in regression tasks, while ChemLLaMA performs better in classification tasks when considering only the model type.

For regression tasks, larger-sized ChemBART models using smaller data sets emerge as one of the best configurations. Yet, ChemLLaMA exhibits a clear proportional relationship between performance and model size, a pattern not observed in the other models. Additionally, ChemLLaMA with larger model sizes outperforms all other foundational model types across various tasks. Notably, ChemLLaMA’s performance in classification tasks also shows a direct correlation with the data size of the pre-training models. When sufficient computational resources and MTR datasets are available, training large-scale ChemLLaMA models with extensive datasets proves to be the most effective strategy for both regression and classification tasks.

\section*{References}
\medskip
{
\small

[1] Deng, J., Yang, Z., Wang, H., Ojima, I., Samaras, D.\ \& Wang, F.\ (2023) A systematic study of key elements underlying molecular property prediction. Nature Communications, {\bf 14}(1), 6395.

[2] David, L., Thakkar, A., Mercado, R.\ \& Engkvist, O.\ (2020) Molecular representations in AI-driven drug discovery: a review and practical guide. {\it J Cheminformatics} {\bf 12}: 56.

[3] Devlin, J., Chang, M. W., Lee, K.\ \& Toutanova, K.\ (2018) Bert: Pre-training of deep bidirectional transformers for language understanding. {\it arXiv:1810.04805.}

[4] Lewis, M., Liu, Y., Goyal, N., Ghazvininejad, M., Mohamed, A., Levy, O., ...\ \& Zettlemoyer, L.\ (2019) Bart: Denoising sequence-to-sequence pre-training for natural language generation, translation, and comprehension. {\it arXiv:1910.13461.}

[5] Touvron, H., Lavril, T., Izacard, G., Martinet, X., Lachaux, M. A., Lacroix, T., ...\ \& Lample, G.\ (2023) Llama: Open and efficient foundation language models. \it arXiv:2302.13971.}

[6] Kim, H., Lee, J., Ahn, S.\ \& Lee, J. R.\ (2021) A merged molecular representation learning for molecular properties prediction with a web-based service. {\it Scientific Reports} {\bf 11}(1), 11028.

[7] Ahmad, W., Simon, E., Chithrananda, S., Grand, G.\ \& Ramsundar, B.\ (2022) Chemberta-2: Towards chemical foundation models. {\it arXiv:2209.01712.}

[8] Irwin, R., Dimitriadis, S., He, J.\ \& Bjerrum, E. J.\ (2022) Chemformer: a pre-trained transformer for computational chemistry. {\it Machine Learning: Science and Technology} {\bf 3}(1), 015022.

[9] Nvidia. (2022). MegaMolBART. Retrieved from {\it https://github.com/NVIDIA/MegaMolBART.}

[10] Hödl, S., Robinson, W., Bachrach, Y., Huck, W.\ \& Kachman, T.\ (2023) Explainability Techniques for Chemical Language Models. {\it arXiv:2305.16192.}

[11] Lang, A. S. I. D., Chong, W. C.\ \& Wörner, J. H.\ (2023) Fine-Tuning ChemBERTa-2 for Aqueous Solubility Prediction. {\it Ann. Chem. Sci. Res}, 4, 1-3.

[12] Liao, C., Yu, Y., Mei, Y.\ \& Wei, Y.\ (2024) From Words to Molecules: A Survey of Large Language Models in Chemistry. {\it arXiv:2402.01439.}

[13] Yu, B., Baker, F. N., Chen, Z., Ning, X.\ \& Sun, H.\ (2024) LlaSMol: Advancing Large Language Models for Chemistry with a Large-Scale, Comprehensive, High-Quality Instruction Tuning Dataset. {\it arXiv:2402.09391.}

[14] Ross, J., Belgodere, B., Chenthamarakshan, V., Padhi, I., Mroueh, Y.\ \& Das, P.\ (2022) Large-scale chemical language representations capture molecular structure and properties. {\it Nature Machine Intelligence} {\bf 4}(12), 1256-1264.

[15] Ramsundar, B., Eastman, P., Walters, P., Pande, V., Leswing, K.\ \& Wu, Z.\ (2019) Deep Learning for the Life Sciences. {\it O'Reilly Media,} {\it https://www.amazon.com/Deep-Learning-Life-Sciences-Microscopy/dp/1492039837.}

[16] Landrum, G.\ (2013) Rdkit documentation. {\it Release} {\bf 1}(1-79), 4.

[17] Li, Z., Wallace, E., Shen, S., Lin, K., Keutzer, K., Klein, D.\ \& Gonzalez, J.\ (2020) Train big, then compress: Rethinking model size for efficient training and inference of transformers. {\it In International Conference on machine learning,} pp.\ 5958-5968. PMLR.

% %%%%%%%%%%%%%%%%%%%%%%%%%%%%%%%%%%%%%%%%%%%%%%%%%%%%%%%%%%%%

% \appendix

% \section{Appendix / supplemental material}

% \begin{figure}
%   \centering
%   % \fbox{\rule[-.5cm]{0cm}{4cm} \rule[-.5cm]{4cm}{0cm}}
%   \includegraphics[width=13.97cm]{ChemLLaMA/appen_whole_reg.png}
%   \label{fig-loss-20m}
%   \caption{Plot of RMSE for all possible model combinations (MTMSDS) from each `best metrics set' of transformer models in the research. Lower values indicate better performance. }
% \end{figure}

% \begin{figure}
%   \centering
%   % \fbox{\rule[-.5cm]{0cm}{4cm} \rule[-.5cm]{4cm}{0cm}}
%   \includegraphics[width=13.97cm]{ChemLLaMA/appen_whole_cls.png}
%   \label{fig-loss-20m}
%   \caption{Plot of RMSE for all possible model combinations (MTMSDS) from each `best metrics set' of transformer models in the research. Lower values indicate better performance. }
% \end{figure}
% % Optionally include supplemental material (complete proofs, additional experiments and plots) in appendix.
% % All such materials \textbf{SHOULD be included in the main submission.}

% %%%%%%%%%%%%%%%%%%%%%%%%%%%%%%%%%%%%%%%%%%%%%%%%%%%%%%%%%%%%
\end{document}